\documentclass[11pt,a4paper]{article}
\usepackage[nohyperref]{acl18-latex/acl2018}
\usepackage{times}
\usepackage{latexsym}
\usepackage{multirow}
\usepackage{array}
\usepackage{amsmath}
\usepackage{amssymb}
\usepackage{graphicx}

\usepackage{url}

\aclfinalcopy 
\def\aclpaperid{xxx} 

\newcommand{\ent}[1]{\textit{#1}}

\newcolumntype{P}[1]{>{\centering\arraybackslash}p{#1}} 
\newcolumntype{M}[1]{>{\centering\arraybackslash}m{#1}} 

\title{Jointly Embedding Entities and Text with Distant Supervision}

\author{Denis Newman-Griffis$^{\clubsuit,\spadesuit}$, Albert M Lai$^{\spadesuit,\blacklozenge}$, \and Eric Fosler-Lussier$^{\clubsuit}$\\
  $^{\clubsuit}$Department of Computer Science and Engineering, The Ohio State University, Columbus, OH\\
  $^{\spadesuit}$Rehabilitation Medicine Department, Clinical Center, National Institutes of Health, Bethesda, MD\\
  $^{\blacklozenge}$Institute for Informatics, Washington University in St. Louis, St. Louis, MO\\
  {\tt \{newman-griffis.1, fosler-lussier.1\}@osu.edu } \ \ \ {\tt amlai@wustl.edu}
}

\date{}

\begin{document}
\maketitle
\begin{abstract}
    Learning representations for knowledge base entities and concepts is becoming
increasingly important for NLP applications.
However, recent entity embedding methods have relied on
structured resources that are expensive to create for new domains and corpora.
We present a distantly-supervised method for jointly
learning embeddings of entities and text from an unnanotated corpus, using
only a list of mappings between entities and surface forms.
We learn embeddings from open-domain and biomedical corpora, and compare
against prior methods that rely on human-annotated text or
large knowledge graph structure.
Our embeddings capture entity similarity and relatedness better than prior
work, both in existing biomedical datasets and a new
Wikipedia-based dataset that we release to the community. Results on analogy
completion and entity sense disambiguation indicate that entities and words
capture complementary information that can be effectively
combined for downstream use.

\end{abstract}

\section{Introduction}

Distributed representations of knowledge base entities and concepts have become
key elements of many recent NLP systems, for applications from document ranking
\cite{Jimeno-Yepes2015} and knowledge base completion \cite{Toutanova2015} to
clinical diagnosis code prediction \cite{Choi2016, Choi2016b}. These works have
taken two broad tacks for the challenge of learning to represent entities, each
of which may have multiple unique surface forms in text. 
Knowledge-based approaches learn entity representations based on the structure
of a large knowledge base, often augmented by annotated text resources
\cite{Yamada2016,Cao2017}. Other methods utilize explicitly annotated
data, and have been more popular in the biomedical domain \cite{Choi2016,
Mencia2016}. Both approaches, however, are often limited by ignoring some or
most of the available textual information. Furthermore, such rich structures
and annotations are lacking for many specialized domains, and can be
prohibitively expensive to obtain.

We propose a fully text-based method for jointly learning representations of
words, the surface forms of entities, and the entities themselves, from an
unannotated text corpus. We use distant supervision from a
\textit{terminology}, which maps entities to known surface forms.  We augment
the well-known log-linear skip-gram model \cite{Mikolov2013a} with additional
term- and entity-based objectives, and evaluate our learned embeddings in both
intrinsic and extrinsic settings.

Our joint embeddings clearly outperform prior entity embedding methods on
similarity and relatedness evaluations. Entity and word embeddings
capture complementary information, yielding improved performance when they
are combined. Analogy completion results further illustrate these
differences, demonstrating that entities capture domain knowledge,
while word embeddings capture morphological and lexical information. Finally,
we see that an oracle combination of entity and text embeddings nearly matches
a state of the art unsupervised method for biomedical word sense
disambiguation that uses complex knowledge-based approaches.
However, our embeddings show a significant
drop in performance compared to prior work in a newswire disambiguation dataset,
indicating that knowledge graph structure contains entity information
that a purely text-based approach does not capture.

\section{Related Work}

Knowledge-based approaches to entity representation are well-studied in recent literature.
Several approaches have learned representations from knowledge graph structure alone \cite{Grover2016,
Yang2016b,Wang2017}.  \newcite{Wang2014b}, \newcite{Yamada2016}, and \newcite{Cao2017} all use a joint embedding method,
learning representations of text from a large corpus and entities from a knowledge graph; however, they rely
on the disambiguated entity annotations in Wikipedia to align their models.  \newcite{Fang2016} investigate
heuristic methods for joint embedding without annotated entity mentions, but still rely on graph
structure for entity training.

The robust terminologies available in the biomedical domain
have been instrumental to several recent annotation--based
approaches.  \newcite{DeVine2014} use string matching heuristics to find possible
occurrences of known biomedical concepts in literature abstracts, and use the sequence of these noisy
concepts (without the document text) as input for skip-gram training.
\newcite{Choi2016CRI} and
\newcite{Choi2016} use sequences of structured medical observations from patients' hospital stays
for context-based learning.  Finally, \newcite{Mencia2016} take documents tagged with Medical Subject Heading
(MeSH) topics, and use their texts to learn representations of the MeSH headers.  These methods are
able to draw on rich structured and semi-structured data from medical databases, but discard important textual
information, and empirically are limited in the scope of the vocabularies they can embed.

\section{Methods}

In order to jointly learn entity and text representations from an unannotated corpus, we use
distant supervision \cite{Mintz2009} based on known {\it terms},
strings which can represent one or more entities.  The mapping between terms and entities
is many-to-many; for example, the same infection can be expressed as ``cold'' or ``acute rhinitis'', but ``cold''
can also describe the temperature or refer to chronic obstructive lung disease.

Mappings between terms and entities are defined by a terminology.\footnote{
    {\it Terminology} is overloaded with both biomedical and lexical senses; we use it
    here strictly to mean a mapping between terms and entities.
}  We extracted terminologies from two well-known knowledge bases:

\textbf{The Unified Medical Language System} (UMLS; \citealp{Bodenreider2004});
we use the mappings between concepts and strings in the MRCONSO table
as our terminology.  This yields 3.5 million entities, represented by
7.6 million strings in total.

\textbf{Wikipedia}; we use page titles and redirects as our terminology.
This yields 9.7 million potential entities (pages), represented by 17.1 million
total strings.
Table~\ref{tbl:terminologies} gives further statistics about the mapping
between entities and surface forms in each of these terminologies.

\begin{table}[t]
    \centering
    {\small
    \begin{tabular}{l|c|c}
        &UMLS&Wikipedia\\
        \hline
        \# entities&3,590,353&9,723,785\\
        \# terms&7,558,254&17,147,756\\
        Max terms&495&7,077\\
        \hline
        \multicolumn{3}{l}{\it \# entities represented by $n$ terms}\\
        \hline
        \ \ $n=1$&1,823,569 (51\%)&6,828,958 (70\%)\\
        \ \ $n=2$&894,932 (25\%)&1,565,109 (16\%)\\
        \ \ $3\leq n\leq10$&831,494 (23\%)&1,143,452 (12\%)\\
        \ \ $n>10$&40,358 (1\%)&186,266 (2\%)\\
        \hline
        \multicolumn{3}{l}{\it \# terms mapping to $n$ entities}\\
        \hline
        \ \ $n=1$&7,473,902 (98\%)&16,127,138 (94\%)\\
        \ \ $n=2$&69,816 (1\%)&958,242 (5\%)\\
        \ \ $3\leq n\leq10$&14,366 ($<1\%$)&62,062 ($<1\%$)\\
        \ \ $n>10$&170 ($\ll1\%$)&15 ($\ll1\%$)\\
    \end{tabular}
    }
    \caption{Statistics of the many-to-many mapping between terms and entities
             in our terminologies, including the maximum \# of terms per
             entity.}
    \label{tbl:terminologies}
\end{table}

While iterating through the training corpus, we identify any exact matches of the terms in our terminologies.\footnote{
    We lowercase and strip special characters and punctuation from both terms and corpus
    text, and then find all exact matches for the terms.
}  We allow for overlapping terms: thus, ``in New
York City'' will include an occurrence of both the terms ``New York'' and ``New York
City.'' Each matched term may refer to one or more entities; we do not use a disambiguation model
in preprocessing, but rather assign
a probability distribution over the possible entities.

\subsection{Model}

We extend the skip-gram model of \newcite{Mikolov2013a},
to jointly learn vector representations of words, terms, and entities from
shared textual contexts.
For a given target word, term, or entity $v$, let $C_v = c_{-k}\dots c_{k}$ be
the observed contexts in a window of $k$ words to the left and right of $v$,
and let $N_v = n_{-k,1}\dots n_{k,d}$ be the $d$ random negative samples for
each context word.
Then, the context-based objective for training $v$ is

\vspace{-0.33cm}
{\small
\begin{equation}
    \label{eq:word-objective}
    O(v,C_v,N_v) = \sum_{c \in C_v}\textrm{log}\sigma(\vec{c}\cdot\vec{v}) + \sum_{n \in N_v}\textrm{log}\sigma(-\vec{n}\cdot\vec{v})
\end{equation}
}

\vspace{-0.33cm}
\noindent where $\sigma$ is the logistic function.

We use a sliding context window to iterate through our corpus.  At each
step, the word $w$ at the center of the window $C_w$ is updated using $O(w,C_w,N_w)$,
where $N_w$ are the randomly-selected negative samples.

As terms are of variable token length, we treat each term $t$ as an atomic unit for training,
and set $C_t$ to be the context words prior to the first token of the term and following the final
token.  Negative samples $N_t$ are sampled independently of $N_w$.

Finally, each term $t$ can represent a set of entities $E_t$.  Vectors for these entities are updated
using the same $C_t$ and $N_t$ from $t$. Since the entities are latent, we weight updates
with uniform probability $|E_t|^{-1}$; attempts to learn this probability did not produce qualitatively
different results from the uniform distribution.
Thus, letting $T$ be the set of terms completed at $w$,
the full objective function to maximize is:

\vspace{-0.33cm}
{\small
\begin{equation}
    \label{eq:objective}
    \begin{split}
        \hat{O} =\ &O(w,C_w,N_w) + \\
            &\sum_{t \in T}\Big[O(t,C_t,N_t) + \sum_{e \in E_t}\frac{1}{|E_t|}O(e,C_t,N_t)\Big]
    \end{split}
\end{equation}
}

\vspace{-0.33cm}
Term and entity
updates are only calculated when the final token of one or more terms is reached; word updates are
applied at each step. To assign more weight to near contexts, we subsample the window
size at each step from $[1,k]$.

\subsection{Training corpora}
\begin{table}[t]
    \centering
    {\small
    \begin{tabular}{l|c|c|c}
        \multicolumn{1}{l}{~}&Pubmed&Wikipedia&Gigaword\\
        \hline
        \# tokens&2.6B&1.9B&4.3B\\
        \# mentions&1.5B&1.4B&3.2B\\
        Avg $CP$&2.54&1.01&1.01\\
        \hline
        \multicolumn{4}{l}{\% of entities by polysemy impact}\\
        \hline
        \ \ $CP\geq1$&99.1\%&98.6\%&98.8\%\\
        \ \ $CP\geq2$&9.3\%&3.5\%&2.2\%\\
        \ \ $CP\geq10$&0.3\%&0\%&$\ll0.1\%$\\
    \end{tabular}
    }
    \caption{Statistics of our embedding training corpora. \# mentions is the
             number of exact matches found for terms in the relevant
             terminology. CP = corpus polysemy of a given entity. B = billion.}
    \label{tbl:corpus-polysemys}
\end{table}

We train embeddings on three corpora.  For our biomedical embeddings, we use
2.6 billion tokens of biomedical abstract texts from the 2016 PubMed baseline (1.5 billion
noisy annotations).
For comparison to previous open-domain work, we use English Wikipedia (5.5
million articles from the 2018-01-20 dump); we also use the Gigaword 5 newswire
corpus \cite{Gigaword5}, which does not have gold entity annotations.

As our model does not include a disambiguation module for handling ambiguous
term mentions, we also calculate the expected effect of polysemous terms on
each entity that we embed using a given corpus. We call this the
entity's \textit{corpus polysemy}, and denote it with $CP(e)$. For entity $e$
with corresponding terms $T_e$, $CP(e)$ is given as
\begin{equation}
    CP(e) = \sum_{t\in T_e}\frac{f(t)}{Z}\textrm{polysemy}(t)
\end{equation}
\noindent where $f(t)$ is the corpus frequency of term $t$, $Z$ is the
frequency of all terms in $T_e$, and polysemy$(t)$ is the number of
entities that $t$ can refer to.

Table~\ref{tbl:corpus-polysemys} breaks down expected polysemy impact for
each corpus. The vast majority of entities experience some polysemy
effect in training, but very few have an average ambiguity per mention
of 50\% or greater. Most entities with high corpus polysemy
are due to a few highly ambiguous generic strings, such as
\textit{combinations} and \textit{unknown}. However, some specific terms
are also high ambiguity: for example, \textit{Washington County} refers to 30
different US counties.

\subsection{Hyperparameters}

For all of our embeddings, we used the following hyperparameter settings:
a context window size of 2, with 5 negative samples per word; initial
learning rate of 0.05 with a linear decay over 10 iterations through the
corpus; minimum frequency for both words and terms of 10, and a subsampling
coefficient for frequent words of 1e-5.

\subsection{Baselines}

We compare the words, terms,\footnote{
    Unknown terms were handled by backing off to words.
} and entities learned in our model
against two prior biomedical embedding methods, using pretrained embeddings
from each. \newcite{DeVine2014} use sequences of automatically identified
ambiguous entities for skip-gram training,
and \newcite{Mencia2016} use texts of documents tagged with MeSH headers to
represent the header codes.  The most recent comparison method for Wikipedia entities is
MPME \cite{Cao2017}, which uses link anchors and graph structure to augment
textual contexts.  We also include skip-gram vectors as a final baseline;
for Pubmed, we use pretrained embeddings with optimized hyperparameters from
\newcite{Chiu2016b}, and we train our own embeddings with word2vec for both
Wikipedia and Gigaword.

\section{Evaluations}
\begin{table}[t]
    \centering
    {\footnotesize
    \begin{tabular}{l|cc|cc}
        &\multicolumn{2}{c|}{Full}&\multicolumn{2}{c}{Filtered}\\
        Method&Sim&Rel&Sim&Rel\\
        \hline
        \multicolumn{5}{l}{\textit{Prior work}}\\
        \hline
        \ \ word2vec         &0.559         &0.496         &&\\
        \ \ DeVine'14        &0.455         &0.422         &0.534         &0.482         \\
        \ \ Mencia'16        &0.565         &0.534         &0.573         &0.536         \\
        \hline
        \multicolumn{5}{l}{\textit{Proposed}}\\
        \hline
        \ \ Word             &0.561         &0.490         &&\\
        \ \ Term             &0.619         &0.557*        &&\\
        \ \ Entity           &0.633*        &0.563*        &0.614*        &0.567*         \\
        \ \ Entity+Word      &0.653*        &0.586*        &0.615*        &\textbf{0.583}*\\
        \ \ \ \ \ \ \ \ \ \ \ \ +Cross&\textbf{0.662}*&\textbf{0.588}*&\textbf{0.622}*&0.573*         \\
        \hline
    \end{tabular}
    }
    \caption{Spearman's $\rho$ for similarity/relatedness predictions in
             UMNSRS.
             Filtered results indicate performance on the shared-vocabulary
             subset.
             *=significantly better ($p<0.05$) than word baseline (full), DeVine et al
             (filtered).}
    \label{tbl:simrel-umnsrs}
\end{table}

Following \newcite{Chiu2016a}, \newcite{Cao2017}, and others, we evaluate our
embeddings on both intrinsic and extrinsic tasks. To evaluate the semantic
organization of the space, we use the standard intrinsic evaluations of
similarity and relatedness and analogy completion. To explore the applicability
of our embeddings to downstream applications, we apply them to named entity
disambiguation. Results and analyses for each experiment are discussed
in the following subsections.

\subsection{Similarity and relatedness}

We evaluate our biomedical embeddings on  the UMNSRS datasets
\cite{Pakhomov2010}, consisting of pairs of UMLS
concepts with judgments of similarity (566 pairs) and relatedness (587 pairs),
as assigned by medical experts.  For evaluating our Wikipedia entity
embeddings, we created WikiSRS, a novel dataset of similarity and relatedness
judgments of paired Wikipedia entities (people, places, and organizations),
as assigned by Amazon Mechanical Turk workers.  We followed the design
procedure of \citet{Pakhomov2010} and produced 688 pairs each of similarity
and relatedness judgments; for further details on our released dataset, please
see the Appendix.

For each labeled entity pair, we calculated the cosine similarity of their
embeddings, and ranked the pairs in order of descending similarity.  We
report Spearman's $\rho$ on these rankings as compared to the ranked human
judgments: Table~\ref{tbl:simrel-umnsrs} shows
results for UMNSRS, and Table~\ref{tbl:simrel-wikisrs} for WikiSRS.

As the dataset includes both string and disambiguated entity forms for each
pair, we evaluate each type of embeddings
learned in our model. Additionally, as words and entities are embedded in
the same space (and thus directly comparable), we experiment with two
methods of combining their information. Entity+Word sums
the cosine similarities calculated between the entity embeddings
and word embeddings for each pair; the Cross setting further adds
comparisons of each entity in the pair to the string form of the other.

\subsubsection{Results}

\begin{table}[t]
    \centering
    {\footnotesize
    \begin{tabular}{l|cc|cc}
        &\multicolumn{2}{c|}{Wikipedia}&\multicolumn{2}{c}{Gigaword}\\
        Method&Sim&Rel&Sim&Rel\\
        \hline
        \multicolumn{5}{l}{\textit{Prior work}}\\
        \hline
        \ \ word2vec         &0.630&0.630&0.624&0.623\\
        \ \ MPME             &0.506&0.567&--&--\\
        \hline
        \multicolumn{5}{l}{\textit{Proposed}}\\
        \hline
        \ \ Word             &0.646         &0.655         &0.615&0.600\\
        \ \ Term             &0.607         &0.667         &0.625&0.673\\
        \ \ Entity           &0.594         &0.648         &0.634&0.686\\
        \ \ Entity+Word      &\textbf{0.718}*&\textbf{0.754}*&0.701*&0.722*\\
        \ \ \ \ \ \ \ \ \ \ \ \ +Cross&0.697*         &0.753*         &0.695*&0.729*\\
        \hline
    \end{tabular}
    }
    \caption{Spearman's $\rho$ for similarity/relatedness predictions in
             WikiSRS, training on two corpora. All Proposed results are
             significantly better than MPME; *=significantly
             better than strongest word-level baseline ($p<0.05$).}
    \label{tbl:simrel-wikisrs}
\end{table}

Our proposed method clearly outperforms prior work and text-based baselines on
both datasets.  Further, we see that the words and entities
learned by our model include complementary information, as combining them
further increases our ranking performance by a large margin.  
As the results on UMNSRS could have been due to our model's ability to embed
many more entities than prior methods, we also filtered the
dataset to the 255 similarity pairs and 260 relatedness pairs that all
evaluated entity-level methods could represent;\footnote{
    For WikiSRS, all methods covered all pairs.
} Table~\ref{tbl:simrel-umnsrs}
shows similar gains on this even footing.
We follow \newcite{Rastogi2015} in
calculating significance, and use their statistics to estimate the minimum
required difference for significant improvements on our datasets.

\begin{table*}[t]
    \centering
    {\small
    \begin{tabular}{c|c|c|c}
        Dataset&Words&Entities&Entity+Word+Cross\\
        \hline
        \multirow{3}{*}{UMNSRS}&Iron/Iron&Iron/Iron&Levaquin/Avelox\\
        &Nausea/Vomiting&Sinemet/Sinemet&Enalapril/Lisinopril\\
        &Lipitor/Zocor&Enalapril/Lisinopril&Carboplatin/Cisplatin\\
        \hline
        \multirow{3}{*}{WikiSRS}&Minas Tirith/Minas Morgul&Real Madrid/FC Barcelona&Ferrari/Lamborghini\\
        &Moscow/Moscow Kremlin&Minas Tirith/Minas Morgul&Moscow/Moscow Kremlin\\
        &Norway/Denmark&Charlize Theron/Screen Actor's Guild&Toshiro Mifune/Akira Kurosawa\\
    \end{tabular}
    }
    \caption{Top 3 pairs in the Relatedness datasets, as ranked by different embedding methods.}
    \label{tbl:simrel-methods}
\end{table*}

In UMNSRS, we found that cosine similarity of entities consistently reflected
human judgments of similarity better than of relatedness; this reflects
previous observations by \newcite{Agirre2009} and \newcite{Muneeb2015}.
Interestingly, we see the opposite behavior in WikiSRS, where relatedness is
captured better than similarity in all settings. In fact, we see a number of
errors of relatedness in WikiSRS predictions, e.g., ``Hammurabi I'' and
``Syria'' are marked highly similar, while the composers ``A.R. Rahman'' and
``John Phillip Sousa'' are marked dis-similar.  MPME embeddings tend towards
over-relatedness as well (e.g., ranking ``Richard Feynman'' and
``Paris-Sorbonne University'' much more highly than gold labels). Despite better
similarity performance, this trend of over-relatedness also holds in biomedical
embeddings: for example, \ent{C0027358} (Narcan) and \ent{C0026549}
(morphine) are consistently marked highly similar across embedding methods,
even though Narcan blocks the effects of opioids like morphine.

\subsubsection{Comparing entities and words}

We observe clear differences in the rankings made by entity vs word
embeddings. As shown in Table~\ref{tbl:simrel-methods}, highly related
entities tend to have high cosine similarity, while word embeddings are more
sensitive to lexical overlap and direct cooccurrence. Combining both sources
often gives the most inuitive results, balancing lexical effects with
relatedness. For example, while the top three pairs by combination in WikiSRS
are likely to co-occur, the top three in UMNSRS are pairs of drug
choices (antibiotics, ACE inhibitors, and chemotherapy drugs, respectively),
only one of which is likely to be prescribed to any given patient at once.

These differences also play out in erroneous predictions. Entity embeddings
often fix the worst misrankings by words: for example, ``Tony Blair'' and
``United Kingdom'' (gold rank: 28) are ranked highly unrelated (position
633) by words, but entities move this pair back up the list (position 86).
However, errors made by entity embeddings are often also made by words:
e.g., \ent{C0011175} (dehydration) and \ent{C0017160} (gastroenteritis) are
erroneously ranked as highly unrelated by both methods. Interestingly, we
find no correlation between the corpus polysemy of entity pairs and
ranking performance, indicating that ambiguity of term mentions
is not a significant confound for this task.

\subsection{Analogy completion}

We use analogy completion to further explore the properties of our joint
embeddings. Given analogy $a:b::c:d$, the task is to guess $d$ given $(a,b,c)$,
typically by choosing the word or entity with highest cosine similarity to
$b - a + c$ \cite{Levy2014}. We report accuracy using the top guess
(ignoring $a,b$, and $c$ as candidates, per \citealp{Linzen2016}).

\subsubsection{Biomedical analogies}

To compare between word and entity representations, we use the entity-level
biomedical dataset BMASS \cite{Newman-Griffis2017}, which includes both entity
and string forms for each analogy. In order to test if words and entities are
capturing complementary information, we also include an oracle evaluation, in
which an analogy is counted as correct if either words or entities produce a
correct response.\footnote{
    We use the Multi-Answer setting for our evaluation (a single $(a,b,c)$
    triple, but a set of correct values for $d$).
} We do not compare against prior biomedical
entity embedding methods on this dataset, due to their limited vocabulary.

\begin{table}[t]
    \centering
    {\small
    \begin{tabular}{l|ccccc}
        Method&B3&H1&C6&L1&L6\\
        \hline
        Words&2.9&0.4&\textbf{7.9}&\textbf{51.5}&\textbf{69.3}\\
        Entities&\textbf{18.3}&\textbf{22.4}&4.5&10.6&10.0\\
        \hline
        Oracle&20.7&22.9&12.1&55.0&70.9\\
    \end{tabular}
    }
    \caption{Accuracy \% on 5 of the relations in BMASS with greatest absolute
             difference in word performance vs entity performance: B3
             (\textit{gene-encodes-product}), H1 (\textit{refers-to}), C6
             (\textit{associated-with}), L1 (\textit{form-of}),
             and L6 (\textit{has-free-acid-or-base-form}). The better of word
             and entity performance is highlighted; all entity vs word
             differences are significant (McNemar's test; $p\ll0.01$).}
    \label{tbl:bmass-results}
\end{table}

Table~\ref{tbl:bmass-results} contrasts the performance of different
jointly-trained representations for five relations with the largest performance
differences from this dataset. For \textit{gene-encodes-product} and
\textit{refers-to}, both of which require structured domain knowledge, entity
embeddings significantly outperform word-level representations. Many of the
errors made by word embeddings in these relations are due to lexical
over-sensitivity: for example, in the renaming analogy \textit{spinal
epidural hematoma:epidural hemorrhage::canis familiaris:\underline{\phantom{dog}}},
words suggest latinate completions such as \textit{latrans} and
\textit{caballus}, while entities capture the correct \ent{C1280551} (dog).
However, on more morphological relations such as
\textit{has-free-acid-or-base-form}, words are by far the better option.

The success of the oracle combination method for entity and word predictions
clearly indicates that not only are words and entities capturing different
knowledge, but that it is complementary. In the majority of the 25 relations
in BMASS, oracle results improved on words and entities alone by at
least 10\% relative. In some cases, as with
\textit{has-free-acid-or-base-form}, one method does most of the heavy
lifting. In several others, including the challenging (and open-ended)
\textit{associated-with}, entities and words capture nearly orthogonal cases,
leading to large jumps in oracle performance.

\subsubsection{General-domain analogies}

No entity-level encyclopedic analogy dataset is available, so we follow
\newcite{Cao2017} in evaluating the effect of joint training on words using the Google analogy
set \cite{Mikolov2013a}.  As shown in Table~\ref{tbl:google-results},
our Wikipedia embeddings roughly match MPME embeddings (which use annotated entity links)
on the semantic portion of the dataset, but our ability to train on unannotated Gigaword
boosts our results on all relations except \textit{city-in-state}.\footnote{
    We failed to precisely replicate the analogy numbers reported by \newcite{Cao2017};
    we attribute this primarily to the different training corpus and slightly different
    preprocessing.
}  Overall, we find that jointly-trained word embeddings split performance
with word-only skipgram training, but that word-only training tends to get
consistently closer to the correct answer. This suggests that
terms and entities may conflict with word-level semantic signals.

\subsection{Entity disambiguation}

Finally, to get a picture of the impact of our embedding method on downstream
applications, we investigated entity disambiguation.\footnote{
    This task is also referred to as entity linking and entity
    sense disambiguation.
} Given a named entity occurrence in context, the task is to assign a canonical
identifier to the entity being referred to: e.g., to mark that ``New York''
refers to the city in the sentence, ``The mayor of New York held a press
conference.'' It bears noting that in unambiguous cases, a terminology alone
is sufficient to link the correct entity: for example, ``Barack Obama'' can only
refer to a single entity, regardless of context. However, many entity strings (e.g.,
``cold'', ``New York'') are ambiguous, necessitating the use of alternate sources
of information such as our embeddings to assign the correct entity.

\subsubsection{Biomedical abstracts}
\label{sssec:msh-wsd}

\begin{table}[t]
    \centering
    {\small
    \begin{tabular}{l|P{0.6cm}P{0.6cm}P{0.6cm}P{0.6cm}P{0.6cm}}
        Method&\scriptsize{Capital (common)}&\scriptsize{Capital (all)}&\scriptsize{Currency}&\scriptsize{City in State}&\scriptsize{Family}\\
        \hline
        word2vec (W)&89.1         &86.0         &15.0         &\textbf{55.5}&\textbf{82.4}\\
        word2vec (G)&90.9         &89.7         &\textbf{18.4}&38.4         &81.0\\
        \hline
        MPME (W)    &83.6         &80.5         &11.9         &50.6         &78.9\\
        \hline                                                
        Proposed (W)&90.1         &78.7         &9.1          &42.5         &75.5\\
        Proposed (G)&\textbf{92.7}&\textbf{92.3}&16.4         &31.3         &81.6\\
    \end{tabular}
    }
    \caption{Analogy completion accuracy \% on the semantic relations in
             the Google analogy dataset. W=Wikipedia, G=Gigaword.}
    \label{tbl:google-results}
\end{table}

We evaluate on the MSH WSD dataset \cite{Jimeno-Yepes2011}, a
benchmark for biomedical word sense disambiguation. MSH WSD consists of mentions
of 203 ambiguous terms in biomedical literature, with over 30,000 total
instances. Each sample is annotated with the set of UMLS entities the term
could refer to. We adopt the unsupervised method of
\newcite{Sabbir2017}, which combines cosine similarity and projection
magnitude of an entity representation $e$ to the averaged word embeddings of its
contexts $C_{avg}$ as follows:
\begin{equation}
    \label{eq:msh-wsd}
    f(e,C_{avg}) = \mathrm{cos}(C_{avg},e)\cdot \frac{||P(C_{avg},e)||}{||e||}
\end{equation}
The entity maximizing this score is predicted.

We compare against concept embeddings learned by \newcite{Sabbir2017}. They used
MetaMap \cite{Aronson2010} with the disambiguation module enabled on a curated
corpus of 5 million Pubmed abstracts to create a UMLS concept cooccurrence
corpus for word2vec training. As shown in Table~\ref{tbl:msh-wsd-results}, our
method lags behind theirs, though it clearly beats both random (49.7\%
accuracy) and majority class (52\%) baselines. In addition, we leverage our
jointly-embedded entities and words by adding in the definition-based model
used by \newcite{Pakhomov2016}, which calculates an entity's embedding as the
average of definitions of its neighbors in the UMLS hierarchy
\cite{McInnes2011}.  We use this alternate entity embedding in
Equation~\ref{eq:msh-wsd} to calculate a second score that we add to the direct
entity embedding score.  This yields a large performance boost of over 6\%
absolute, indicating that using entities and words together makes up much of
the gap between our distantly supervised embeddings and the external resources
used by \newcite{Sabbir2017}. Using the definition-based method alone with our
jointly-embedded words, we see a significant increase over \newcite{Pakhomov2016},
indicating the benefits of joint training. However, the combined
entity and definition model still yields a significantly different 2\% boost
in accuracy over definitions alone. Finally, we evaluate an oracle combination that
reports correct if either entity or definition embeddings achieve the correct
result; as shown in the last row of Table~\ref{tbl:msh-wsd-results}, this
combination outperforms the entity-only method of \newcite{Sabbir2017}, and
approaches their state-of-the-art result that combines entity embeddings
with a knowledge-based approach from the structure of the UMLS.

\begin{table}[t]
    \centering
    {\small
    \begin{tabular}{l|c}
        Method&Accuracy \%\\
        \hline
        \multicolumn{2}{l}{\it Baselines}\\
        \hline
        \ \ \newcite{Sabbir2017} (entities; +MetaMap)&89.3\\
        \ \ \newcite{Sabbir2017} (+MetaMap, UMLS)&\textbf{92.2}\\
        \ \ \newcite{Pakhomov2016} (words)&77.7\\
        \hline
        \multicolumn{2}{l}{\it Proposed}\\
        \hline
        \ \ Entities&76.4\\
        \ \ Definitions (joint words)&80.8\\
        \ \ Entities+Definitions&82.7\\
        \ \ Oracle (Entities|Definitions)&90.9\\
    \end{tabular}
    }
    \caption{MSH WSD disambiguation accuracy. Definitions is comparable to
             \newcite{Pakhomov2016}, using jointly-embedded words. All
             differences are significant (McNemar's test, $p\ll0.01$).}
    \label{tbl:msh-wsd-results}
\end{table}

Specific errors shed more light on these differences. The definition-based
method performs better in many cases where the surface form is a common word,
such as {\it coffee} (68\% definition accuracy vs 28\% entity accuracy) and
{\it iris} (93\% definition accuracy vs 35\% entity accuracy). Entities
outperform on some more technical cases, such as {\it potassium} (74\% entity
accuracy vs 49\% definition accuracy). Combining both approaches in the joint
model recovers performance on several cases of low entity accuracy; for example,
joint accuracy on {\it coffee} is 68\%, and on {\it lupus} (53\% entity accuracy),
joint performance is 60\%.

\subsubsection{Newswire entities}

AIDA \cite{Hoffart2011} is a standard dataset for entity linking in newswire,
consisting of approximately 30,000 entities linked to Wikipedia page IDs.
To reduce the search space, \newcite{Pershina2015} provided a set of candidate
entities for each mention, which we use for our experiments.
The MPME model of \newcite{Cao2017} achieves near state-of-the-art performance
accuracy on AIDA with this candidate set, using the mention sense distributions
and full document context included in the model. As our embeddings are trained
without explicit entity annotations, we instead use the same cosine similarity
and projection model discussed in Section~\ref{sssec:msh-wsd} for this task.
In contrast to our results on the biomedical data, we see performance far below
the baseline on these data, as shown in Table~\ref{tbl:aida-results}.

\begin{table}[t]
    \centering
    {\small
    \begin{tabular}{l|c}
        Method&Accuracy \%\\
        \hline
        MPME (entities; +graph structure)&\textbf{89.0}\\
        \hline
        Wikipedia&40.9\\
        Wikipedia + mentions&44.6\\
        Gigaword&58.0\\
        Gigaword + mentions&63.9\\
    \end{tabular}
    }
    \caption{AIDA linking accuracy, using entity embeddings trained
             on Wikipedia and Gigaword. All differences are significant
             (McNemar's test, $p\ll0.01$).}
    \label{tbl:aida-results}
\end{table}

\begin{table*}[t]
    \centering
    {\footnotesize
    \begin{tabular}{M{1.8cm}|M{2.0cm}|M{2.0cm}|M{3.8cm}|M{3.8cm}}
        Entity&Words&Terms&Entities&Joint\\
        \hline
        \multirow{5}{1.8cm}{\centering C0009443 (common cold)}
            &k(+)-grown&cold&C0041912 (upper respiratory infections)&C0041912 (upper respiratory infections)\\
            &legionella-contaminated&short periods&C0234192 (cold sensation)&C0234192 (cold sensation)\\
            &hyperinflating&changed&C0719425 (``Cold'' pharmaceutical brand)&C0719425 (``Cold'' pharmaceutical brand)\vspace{0.2cm}\\
        \hline
        \multirow{6}{1.8cm}{\centering C0242797 (home health aides)}
            &\vspace{0.2cm}homemaker-home&\vspace{0.2cm}home health aide&\vspace{0.2cm}C1553498 (home health encounter)&\vspace{0.2cm}home health aide\\
            &voluntary-sector&home health aides&C0019855 (home care services)&home health aides\\
            &health/social&home health&C1317851 (home health care specialty)&C1553498 (home health encounter)\\
    \end{tabular}
    }
    \caption{Top 3 nearest neighbors to two UMLS entities, using
             words, terms, entities, or all three.}
    \label{tbl:poly-neighbors}
\end{table*}

However, we improve this performance slightly by multiplying by the
similarity between the entity embedding and the average word embedding of
the mention itself; this gives us roughly a further 4\% accuracy for both
Wikipedia and Gigaword embeddings. Using the surface form recovers several
cases where entities alone yield unlikely options, e.g. Roman-era Britain
instead of the United Kingdom for {\it Britain}. However, it also introduces
lexical errors: for example, {\it British}
\begin{figure}[b!]
    \centering
    \includegraphics[width=0.5\textwidth]{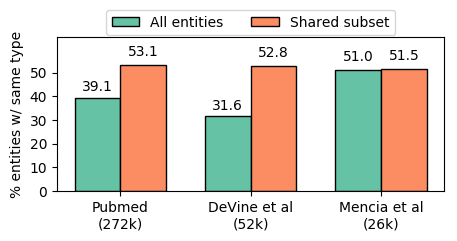}
    \vspace{-0.9cm}
    \caption{Percentage of UMLS entities whose nearest neighbor shares a
             semantic type, with no vocabulary restriction (vocab size in
             parentheses) and in a shared vocabulary subset.}
    \label{fig:neighbor-sty}
\end{figure}

in several cases refers to the United Kingdom, but the British people are
often selected instead. We note that this extra
score actually hurts performance on MSH WSD, where the terms are curated to be
highly ambiguous, in contrast to the shorter contexts and
clearer terms used in AIDA.

Two other issues bear consideration in this evaluation. Prior approaches to
the AIDA dataset, including MPME, make use of the global context of entity
mentions within a document to improve predictions; by using local context only,
we observe some inconsistent predictions, such as selecting the cricket world
cup instead of the FIFA competition for {\it world cup}, in a document discussing
football. Additionally, in contrast to the MSH WSD dataset, many instances in AIDA
have several highly-related candidates that introduce some confusion in our results.
For example, {\it Ireland} could refer to the United Kingdom of Great Britain and
Ireland, the island of Ireland, or the Republic of Ireland. As our embedding training
does not include gold entity links, cases like this are often errors in our predictions.

\section{Analysis of joint embeddings}

To get a more detailed picture of our joint embedding
space, we investigate nearest neighbors for each point by cosine similarity.
As entities in the UMLS are assigned one or more of over 120 semantic types, we first
examine how intermixed these types are in our biomedical embeddings.
Figure~\ref{fig:neighbor-sty} shows how often an entity's nearest neighbor
shares at least one semantic type with it, across the three
biomedical embedding methods we evaluated. As each set of embeddings has a
different vocabulary, we also restrict to the entities that all three can
embed (approximately 11,000).

We see that our method puts entities of the same type together nearly 40\%
of the time, despite embedding over 270 thousand entities. On an even footing,
our method puts types together significantly more often \newcite{Mencia2016}
(McNemar's; $p<0.05$), and equivalently with \newcite{DeVine2014},
despite using less entity-level information in training.
Within our embeddings, major biological types such as bacteria, eukaryotes,
mammals, and viruses all have more than 60\% of neighbors with the same type,
while less structured clinical types such as Clinical Attribute and Daily or
Recreational Activity are in the 10-20\% range. Corpus polysemy does not appear
to have any effect on this type matching (mean polysemy of 1.5 for both matched
and non-matched entities).

Expanding to include the words and terms in the joint embedding space, however,
we see definite qualitative effects of corpus polysemy on entity nearest neighbors.
Table~\ref{tbl:poly-neighbors} gives nearest word, term, entity, and joint
neighbors to two biomedical entities: \ent{C0009443} (the common cold; $CP=6.71$)
and \ent{C0242797} (home health aides; $CP=1$). For the more polysemous
\ent{C0009443}, where 95\% of its mentions are of the word ``cold'' (polysemy=7),
word-level neighbors are mostly nonsensical, while term neighbors are more
logical, and entity neighbors reflect different senses of ``cold''. By contrast,
the non-polysemous \ent{C0242797}, which is represented by 14 different
unambiguous strings, words, terms, and entities are all very clearly in line
with the theme of home health aides. Notably, the common and unambiguous terms
for \ent{C0242797} are its nearest neighbors out of all points, while only two
of the top 10 neighbors to \ent{C0009443} are terms.

\section{Discussion}

\newcite{Faruqui2016} observe that similarity and relatedness are not clearly
distinguished in semantic embedding evaluations, and that it is unclear exactly
how vector-space models should capture them.  We see more evidence of this,
as cosine similarity seems to be capturing a mix of
the two properties in our data.  This mix is clearly informative, but it
empirically favors
relatedness judgments, and cosine similarity is insufficient to separate
the two properties.

Corpus polysemy plays a qualitative role in our embedding model, but
less of a quantitative one. It does not correlate with similarity
and relatedness judgments or entity disambiguation decisions, but it clearly
affects the organization of the embedding space, by embedding
entities with high corpus polysemy in less coherent areas than those with low
polysemy. \newcite{Linzen2016} points out that for analogy completion, local
neighborhood structure can interfere with standard methods; how this
neighborhood structure affects predictions in more complex
tasks is an open question.

Overall, we find two main advantages to our model over prior work. First,
by only using a terminology and an unannotated corpus, we are able to learn
entity embeddings from larger and more diverse data; for example, embeddings
learned from Gigaword (which has no entity annotations) outperform embeddings
learned on Wikipedia in most of our experiments. Second, by embedding entities
and text into a joint space, we are able to leverage complementary information
to get higher performance in both intrinsic and extrinsic tasks; an oracle
model nearly matches
a state-of-the-art ensemble vector and knowledge-based model for biomedical
word sense disambiguation. However,
our other entity disambiguation results demonstrate that there is additional
entity-level information that we are not yet capturing. In particular, it
is unclear whether our low performance on disambiguating newswire entities
is due to a disambiguation model mismatch, a lack of information in our
embeddings, or a combination of both.

\section{Conclusions}

We present a method for jointly learning embeddings of entities and text from
an arbitrary unannotated corpus, using only a terminology for distant supervision. 
Our learned embeddings better capture both biomedical and encyclopedic similarity and
relatedness than prior methods, and approach state-of-the-art performance for
unsupervised biomedical word sense disambiguation. Furthermore, entities and
words learned jointly with our model capture complementary information, and
combining them improves performance in all of our evaluations.
We make an implementation of our method available at
{\tt github.com/OSU-slatelab/JET}, along with
the source code used for our evaluations and our pretrained entity
embeddings.
Our novel Wikipedia
similarity and relatedness datasets are available at the same source.

\section*{Acknowledgments}

We would like to thank Chaitanya Shivade for helpful discussions, and all
of our anonymous reviewers for their invaluable advice.
This research was supported in part by the Intramural Research Program of the
National Institutes of Health, Clinical Research Center and through an
Inter-Agency Agreement with the US Social Security Administration.

\bibliographystyle{acl18-latex/acl_natbib}
\bibliography{references.clean}

\clearpage
\appendix

\section{WikiSRS construction details}
\label{app:wikisrs}

We followed a similar process to \newcite{Pakhomov2010} in selecting the entity pairs to be used in our dataset.
We first filtered the full list of Wikipedia pages to the subset that we learned embeddings for, and then used
the entity types assigned to these pages in YAGO \cite{YAGO3} to restrict to only entities labeled with WordNet
types organization or person, or with the YAGO type geoEntity.  For each pairing of these categories 
(Organization-Organization, Organization-Place, Organization-Person, Place-Place, Place-Person, and Person-Person),
we manually selected 30 pairs of entities for each of the following relatedness categories: Completely Unrelated,
Somewhat Unrelated, Somewhat Related, and Highly Related.  These produced the list of 720 entity pairs we used
for our Mechanical Turk surveys.

We augmented each survey of 30 questions with 4 manually-created validation pairs using common entities (e.g.,
London, New York), each of which was categorized as Highly Related or Completely Unrelated.  We included these
validation questions at random indices in our surveys.  To evaluate if participants were reading the questions,
we binned their ratings on these validation questions into 0-25 (Completely Unrelated), 26-50 (Somewhat Unrelated),
51-75 (Somewhat Related), and 76-100 (Highly Related).  If a participant's ratings disagreed with ours on multiple
validation questions, we discarded their data (we allowed disagreement on a single question, as some validation
questions had high variance in responses among reliable annotators).

We recruited 6 participants for each survey, for a total of 34 unique participants across the 48 HITs.  Participants 
were presented with a message describing the survey and stating that by clicking the button at the bottom of the message
to begin the survey, they were providing informed consent to participate.  Identifying participant data was not collected,
and we used only the anonymous worker IDs provided by the Mechanical Turk interface to collate our data and remunerate
workers.  Participants were asked optional demographic questions about their age bracket and native language at the end
of the survey; we did not end up using age information, but filtered our participants for those that self-reported English
reading proficiency.  The majority responded to a single HIT, while 3 completed more than 20.  We discarded all submissions
from 3 participants, as they did not report English reading proficiency (1) or did not satisfy the validation questions (2).
All participants were paid state minimum wage at the time of the study for their time, regardless of whether they answered
demographic questions or if we used their data in the final sample.  Collection of this data was approved under Ohio State
University IRB protocol 2017E0050.

To generate the final dataset, we assessed each participant's responses to the validation questions in each survey.  We kept
surveys for which we had at least 4 participants with satisfactory answers to the validation questions; this resulted in
discarding 1 of the 24 HITs for each task.  Due to 2 repeated pairs, this gave us final dataset sizes of 688 pairs for each
of similarity and relatedness, 658 of which were shared between the tasks.

\begin{table}[t]
    \centering
    \begin{tabular}{c|cc|cc}
        \multirow{2}{*}{\# of raters}&\multicolumn{2}{c|}{Similarity}&\multicolumn{2}{c}{Relatedness}\\
        &ICC&\# pairs&ICC&\# pairs\\
        \hline
        4&0.531&419&0.467&180\\
        5&0.520&267&0.540&207\\
        6&     &  &0.560&299\\
        $>6$&--&2&--&2\\
        \hline
        Total&&688&&688\\
    \end{tabular}
    \caption{The intraclass correlation coefficient (ICC) among Amazon Mechanical Turk worker judgments of similarity and
             relatedness of pairs of Wikipedia entities.  As ICC requires a fixed number of raters, but we had variable
             numbers of responses to each HIT, we break down the datasets by the number of workers who rated each item.}
    \label{tbl:wikisrs-icc}
\end{table}

Following \newcite{Pakhomov2010}, we assessed inter-annotator agreement using the intraclass correlation coefficient (ICC).
Table~\ref{tbl:wikisrs-icc} gives the values for our datasets. The numbers reported are within the moderate range, and they
correspond to the ICC numbers reported by Pakhomov et al.\ on the UMNSRS datasets.

The source code of our Mechanical Turk interface and data files used to generate the tasks are available
at {\tt github.com/OSU-slatelab/WikiSRS}.

\end{document}